\documentclass[[journal,11pt,onecolumn,final]{IEEEtran}
\IEEEoverridecommandlockouts
\usepackage{url}
\usepackage{subfig}
\usepackage{cite}
\usepackage{amsmath,amssymb,amsfonts}
\usepackage{graphicx}
\usepackage{textcomp}
\usepackage{adjustbox}
\usepackage{xcolor}
\usepackage[textsize=tiny,textwidth=1.4cm]{todonotes}
\usepackage{mathtools,lipsum, nccmath,cuted}
\usepackage{siunitx}

\usepackage{comment}
\usepackage{makecell}
\usepackage{algpseudocode}
\usepackage{gensymb}
\usepackage{ulem}
\usepackage{colortbl}
\usepackage{hyperref}
\usepackage{enumitem}
\usepackage{booktabs} 
\definecolor{Gray}{gray}{0.925}
\usepackage[textsize=tiny]{todonotes}

\graphicspath{ {figures/} }
\def\BibTeX{{\rm B\kern-.05em{\sc i\kern-.025em b}\kern-.08em
    T\kern-.1667em\lower.7ex\hbox{E}\kern-.125emX}}
    
\usepackage{pgfplots}
\usepgfplotslibrary{colorbrewer}
\pgfplotsset{compat = 1.14, cycle list/Set1-8} 
\usetikzlibrary{pgfplots.statistics, pgfplots.colorbrewer,pgfplots.dateplot,
        shadows,
        matrix} 

\usepackage{pgfplotstable}
\usepackage{filecontents}
\usepackage{graphicx}
\pgfplotsset{compat=1.14}

\definecolor{blueLine}{RGB}{57,106,177}
\definecolor{blueFill}{RGB}{114,147,203}
\definecolor{redLine}{RGB}{204,37,41}
\definecolor{greenline}{RGB}{0,250,0}
\definecolor{blackLine}{RGB}{0,0,0}
\definecolor{goldLine}{RGB}{160,82,45}

\usepackage[font=footnotesize]{caption}
\begin{document}
\title{Vehicle lateral control using Machine Learning for automated vehicle guidance }

\author{\IEEEauthorblockN{Akash Fogla*, Kanish Kumar, Sunnay Saurav, Bishnu ramanujan }\\
\IEEEauthorblockA{
\textit{*Department of Mathematics, Georgia State University  }
}
}

\maketitle

\section{\textbf{\textit{ABSTRACT}}}
\textit{ Uncertainty in decision-making is crucial in the machine learning model used for a safety-critical system that operates in the real world. Therefore, it is important to handle uncertainty in a graceful manner for the safe operation of the CPS. In this work, we design a vehicle's lateral controller using a machine-learning model. To this end, we train a random forest model that is an ensemble model and a deep neural network model. Due to the ensemble in the random forest model, we can predict the confidence/uncertainty in the prediction. We train our controller on data generated from running the car on one track in the simulator and tested it on other tracks. Due to prediction in confidence, we could decide when the controller is less confident in prediction and takes control if needed. We have two results to share: first, even on a very small number of labeled data, a very good generalization capability of the random forest-based regressor in comparison with a deep neural network and accordingly random forest controller can drive on another similar track, where the deep neural network-based model fails to drive, and second confidence in predictions in random forest controller makes it possible to let us know when the controller is not confident in  prediction and likely to fail. By creating a threshold, it was possible to take control when the controller is not safe and that is missing in a deep neural network-based controller. 
}
\subsection*{Keywords}
\textbf{\textit{Random forest, Deep neural networks, Torcs, surrogate modeling, Electronics Control Unit}}

\section{Introduction}
\label{sec:introduction}
A self-driving car has many machine learning and AI-based components involved that act as subsystems in a complex driving system like a perception system, obstacle detection system, etc. However, there are very few examples of an end-to-end driving system\cite{bojarski2016end}  by using ML and AI models. ML and AI have shown extraordinary success in the field of controls\cite{abbeel2010autonomous,vardhan2021rare}, prediction or modeling of complex behavior like cancer detection, stock market prediction, etc \cite{al2019comparative,ghazanfar2017using}, to design automation \cite{vardhan2022deepal}.  In this work , we take two famous and extensively used machine learning model and train it to control the vehicle lateral control. The goal of this study is to understand the comparison between different trained model when used in controlling a system  when an adequate amount of data is available. This comparison study can be used as a baseline and can provide guidelines for AI based control designer an idea about selection of AI-ML model and their respective strength and weakness. For vehicle lateral control, we utilize the open-source car racing simulator called TORCS\cite{wymann2000torcs}. On a given set speed , the data is generated using an traditional PID controller designed by experimenter. The PID controller produces brake value, acceleration and steering to control and drive the car on a given track. During training , for lateral control, we are only interested in steering value and brake and acceleration value is taken using the PID controller. The collected data is distance measured by LIDAR sensor suite and the steering value. For training the controller ,we select decision tree based ML model called Random forest\cite{breiman2001random} and Deep neural network model\cite{ivakhnenko1968group,fukushima1988neocognitron}. Both are trained on same data set and tested in same scenario (track and velocity). 
Our experimentation shows following :  
 \begin{enumerate}
     \item  Random forest-based controller provides better generalization capability than Deep neural network when we done have access of very large data set.
     \item When both trained model is deployed on another track, the random forest-based controller was able to complete the track without crashing, while the deep neural network-based controller  failed to complete the track.
     \item The random forest controller can quantify uncertainty in prediction that can help to decide the failure probability of the controller and override the AI control command if needed.  
 \end{enumerate}

The rest of the paper is organised as follow. Section \ref{sec:methods} formulate the problem and provide background, approach and training details of controller. Section \ref{sec:experimentalResults} provides the experimental results of the experimentation. The related work is discussed in section \ref{sec:relatedWorks} and at the end we produce our conclusion and future direction of research\ref{sec:conclusionFutureWork}.

\section{Problem Formulation and Approach}
\label{sec:methods}
\subsection{Problem Formulation}
 For formalization of the problem, we first introduce some notations. Let $x$ is the state of the vehicle that is estimated by the sensors attached to the car and $x$ is sampled from an unknown distribution $P_x$ which we neither control nor know but can sample from it. The goal of controller design is to learn a nonlinear function $f$ such that it maps each $x$ to the control action $c$.
 $$f: x \mapsto c \;  where\; x \sim P_x$$
The choice of $f$ that is an instance of a class of architecture ($\mathcal{A}$) affects the quality of the control action. In this context, we  chose two famous or extensively used architectures called random forest and Deep neural networks. The choice of model architecture also affects the training process and the parameters to select. In this context, we frame this modeling problem as a supervised learning problem where it is possible to get a labeled dataset that can be used for training. The collection of labeled sets of $<x_i,c_i>$ represents our data $D$. This data set is further split into training and test dataset called $D_{train}$ and $D_{test}$. Then the learning goal is to find the parameters (in the case of neural networks or feature decision graphs (in the case of the random forest)  that can represent the training data the most and work well on test data as well. 

\subsection{Approach}
In this section, we discuss a little background literature on our experimentation components, the approach of the experimentation, model training, and related details. 

\textbf{Torcs simulator} TORCS (The Open Racing Car Simulator) is an open-source 3D car racing simulator that has the ability to  drive manually using peripheral devices or by design controller- AI-based or traditional controller. The key components of vehicle dynamics, including mass, rotational inertia, collision, the mechanics of suspensions, links, differentials, friction, and aerodynamics, may all be accurately simulated by it. Using a temporal discretization level of $0.002$ seconds, Euler integration of differential equations is used to simplify and carry out physics simulation. TORCS offers a large variety of tracks and cars as assets. 
\\

\begin{figure}[tbhp]
\centering
   \includegraphics[width=1.0
   \linewidth]{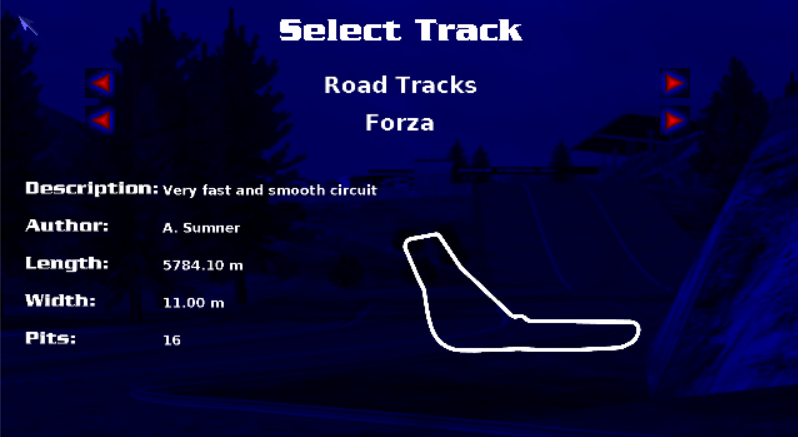}
   \caption{track: Forza}
   \label{fig:forza}
\end{figure}

\textbf{Data generation} For data generation purposes, we used the TORCS' asset car on the road track called 'Forza' (refer to fig \ref{fig:forza}). For sensing the environment, we deployed multiple LIDAR sensors attached to the car that can sense the track and the wall. For data generation purposes, we designed a tuned PID controller that can drive the car successfully on a given speed on the track Forza. Our data set consists of input as a set of distances measured by various LIDAR sensors and output as the steering value. We ran multiple rounds on this track and collected this data. This labeled data set is used for training our models. 
Both models (random forest and deep neural network) can be used for both classification and regression tasks. In our case, we want to use it as a regressor that can predict the steering value on the given LIDAR sensor data. 
\\

\textbf{Random forest model, parameters and training }
Random forest is a learning model that is an ensemble model of various decision trees. For regression tasks, the mean prediction of the individual decision trees is the estimated prediction. Random forests do various other tweaks in the vanilla decision tree for taking care of the habit of over-fitting of decision trees' on their training set.

In particular, decision trees are grown very deep when trained on data and consequently tend to learn highly nonlinear patterns but over-fit, i.e. have low bias, but very high variance. A random forest is an approach of averaging multiple decision trees, where each decision tree is trained on a different sub-sample of the training data set, with the goal of reducing the variance in prediction. 

For training the decision tree in a random forest we apply the technique of bootstrap aggregating, or bagging, to each decision tree learner. On a given input data set $D=\{X,Y\}$ where $X$ is LIDAR's measurement of obstacle distance; $X= \{x_1, x_2,..., x_n\}$ and steering output $Y=\{y_1,y_2,...,y_n\}$, the bagging process involves repeatedly sampling a subset of data with $B$ data points from the data set $D$ with replacement and train each decision tree on this selected samples. If $n$ is the number of trees in the random forest the process involves as below: 
for i=1,...,n:
\begin{enumerate}
    \item Sample with replacement $B$ training examples from $D$; call it $D_i$.
    \item train the $i^{th}$ decision tree $T_i$ on data set $D_i$.   
\end{enumerate}
For prediction on unseen data $x^{'}$, the steering output is estimated by estimating the mean of prediction from each decision tree. 
$$\hat{s}= \frac{1}{n} \sum_{i=1}^{i=n} T_i(x^{'}) $$
Additionally,  the uncertainty in the prediction can be estimated  by calculating the standard deviation of the predictions from all the individual regression trees on $x^{'}$:
$$\sigma= \sqrt{\frac{\sum_{i=1}^{i=n} (T_i(x^{'})-\hat{s})^2}{n-1}} $$
The prediction value of steering $\hat{s}$ is used for steering the vehicle and the standard deviation and other related statistics -like Coefficient of Variance (CoV). We select $n=100$ i.e. total 100 number of decision trees.  The other hyper-parameters are 'splitting criteria= \textit{squared error}', minimum samples for split=2, the maximum number of features=1.0, and minimum impurity decrease=$0.001$. 
\\
\textbf{Deep Neural Network model, parameters and training}
DNN are computing functions inspired by the biological neural networks of animal brains. These systems learn to do a task by observing examples that are already done/labeled. It has wide ranging application from image recognition\cite{}, engineering design\cite{vardhan2021machine,vardhan2022deepal}, control design, anomaly detection\cite{vardhan},etc. A DNN is based on a layered network of smaller computational units called artificial neurons. These neurons are in multiple layers  and fully connected between the input and output layers. A fully connected feed-forward neural network is used in this experiment that is defined by its architecture $a=(L, N,\delta)$, where $L$ is the number of layers in the networks $L \in N$, $N$ is the set that represents the number of neurons in each layer represented by $N_l$ where $l \in [L-1]$, $\delta$ is the activation function $\delta: R \mapsto R$. For this experiment, we used a fully connected feed-forward 6 layers neural network with $\{256,128,64,32,16\}$ neurons in hidden layers and $1$ neuron in the output layer. All hidden neurons have ReLU activation units and the output neuron is the linear activation unit.  The loss function, in this case, is the mean square error and the weight initialization is Xavier normal. The training is done for $500$ epochs by splitting the training and validation data into $90:10$ and early stopping is used to stop over-fitting.  
\begin{figure}[tbhp]
\centering
   \includegraphics[width=1.0
   \linewidth]{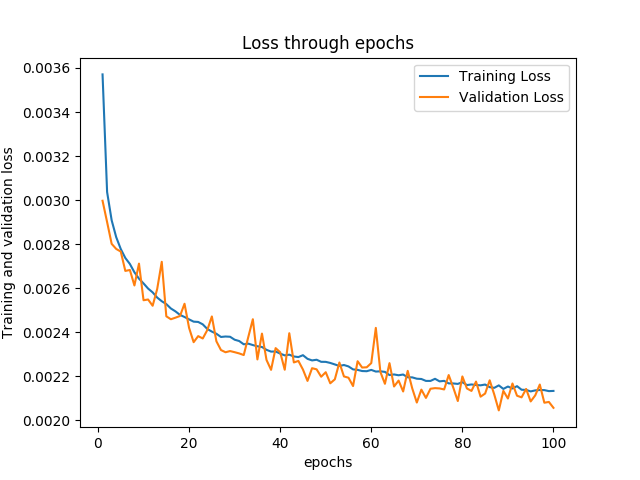}
   \caption{Training and validation loss for deep neural network training}
   \label{fig:nn_train}
\end{figure}

\section{Experimental Results}
\label{sec:experimentalResults}
\begin{figure}[tbhp]
\centering
   \includegraphics[width=1.0
   \linewidth]{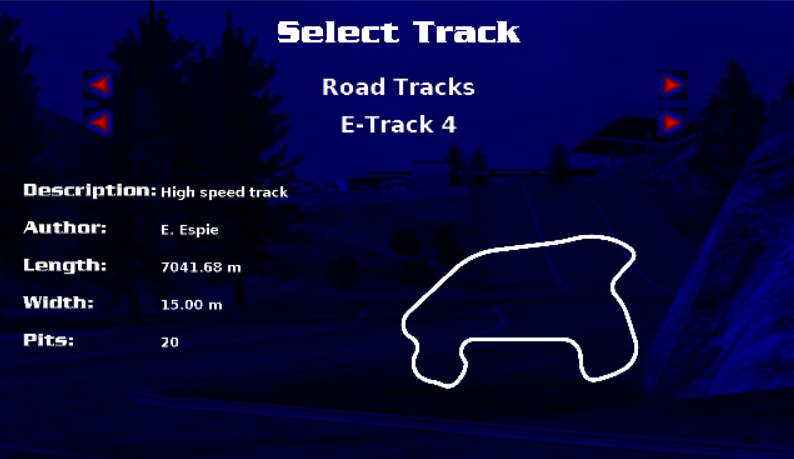}
   \caption{track: E-track 4}
   \label{fig:et4}
\end{figure}

Once both models are trained, these models are tested on a similar but another track called 'E-Track 4' (refer to figure \ref{fig:et4}). The selection of the track is based on twists and turns and complexity in the track profile.  We set the target speed of the car to 60 miles/hrs. For the first experiment, we used both controllers to complete the track without any manual interventions.  The task was to finish one complete lap of this track. One complete lap of this track is 7.041 Km long with a track width of 15 meters.  The random forest controller was able to complete the task without any crash or off-the-track navigation, while the deep neural network-based controller failed to complete the track and crashed multiple times and could not complete even $10\%$ of the track without a crash.  

\begin{figure*}[h!]
\centering
   \includegraphics[width=1.0
   \linewidth]{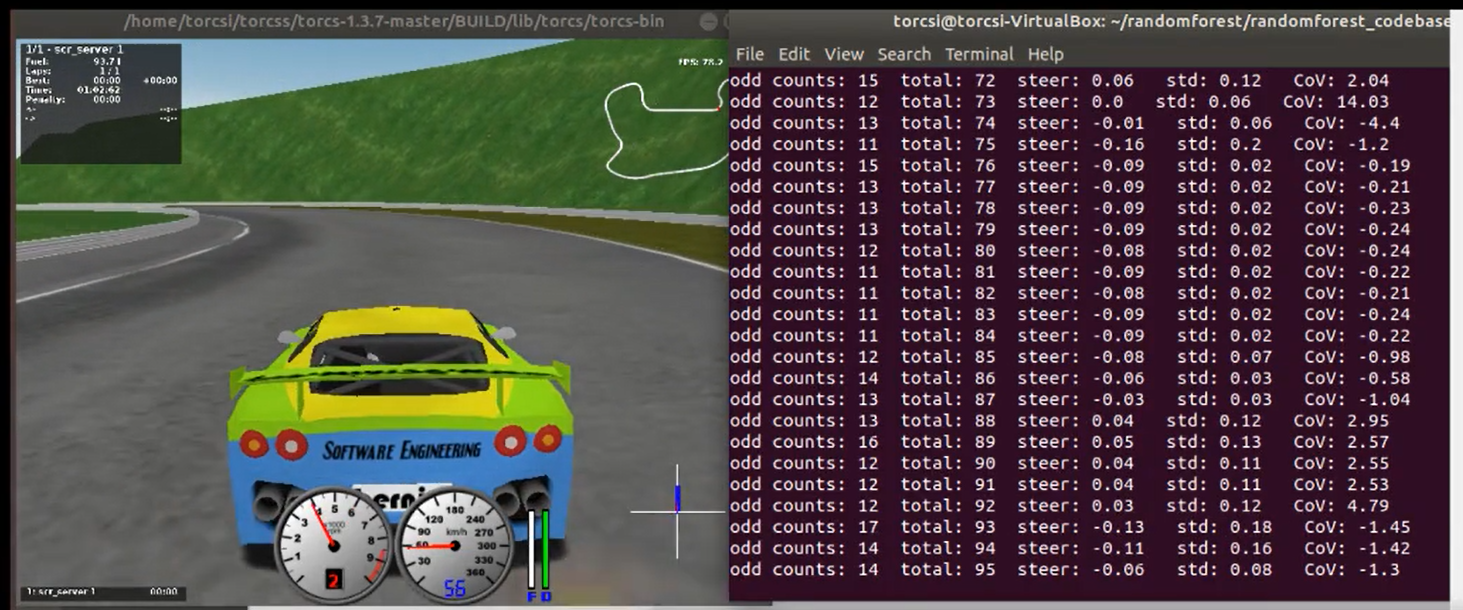}
   \caption{RF controller in Torcs and related statistics, the statistics of our interest are: \textit{odd counts} -number of decision trees out of 2 standard deviations from the mean, \textit{total}- total number of odd counts during whole simulation, \textit{steer}- steering value predicted by regressor, \textit{std}- standard deviation of prediction, \textit{CoV}- coefficient of variance, that explain the spread of the distribution. }
   \label{fig:drive}
\end{figure*}

In the second experiment, we let the experimenter intervene whenever the random forest controller gives a prediction with a high coefficient of variance. A high coefficient of variance (CoV) reflects less confidence in prediction and consequently, its prediction cannot rely upon. In such cases, the control action is shifted to manual control. We observed that the trained random forest controller produced high CoV in scenarios which was very complex and never encountered during training and also when we tried to run it on a significantly different track. This feature is missing with the Deep Neural network controller. A snapshot of the random forest controller driving the car on the test track is shown in figure \ref{fig:drive}. 

\section{Related Work}
\label{sec:relatedWorks}
The dominant approach to solving engineering problems are model-driven, like model-based design\cite{neema2019web}, model-based control\cite{brosilow2002techniques}, and model-based optimization\cite{vardhan2019modeling,neema2019design} etc. With the discovery of data-driven models especially deep neural networks the domain of model-based engineering is revolutionizing.  In recent times these data-driven models have been used and have worked well in image classification (AlexNet\cite{krizhevsky2017imagenet} ), engineering design\cite{vardhan2022data,moosavi2020role,vardhan2023search,volk2020biosystems}, autonomous driving car\cite{bojarski2016end,badue2021self,ni2020survey,shalev2017formal}, radiology \cite{gross1990neural}, human genome\cite{sundaram2018predicting}, and many more for developing state of the art control\cite{vardhan2022reduced,wu2019machine,tahsien2020machine} and prediction systems\cite{vardhan2022deep}.

According to research\cite{momtaz2018rate}, self-driving cars could potentially save up to $35,000$ lives annually in the United States alone by reducing traffic deaths by up to $99$ percent.
$93$ percent of car accidents are the result of driver mistake, according to information from 2007 US report\cite{facts2007us}, and rookie drivers—who seldom have more than $30$ hours of driving experience before receiving a license—are over-represented in fatal car accidents.
Moreover, crashes among senior drivers rise at the same time because they have a diminished capacity to assess their surroundings for unforeseen dangers, according to \cite{momtaz2018rate}.
Self-driving cars will have been trained using machine learning data-sets made up of a wide range of driving circumstances and driver behaviors for the equivalent of hundreds of hours behind the wheel before they hit the roads\cite{bojarski2016end}. The practical utility of AI-ML model is enormous \cite{vardhan2023fusion,dreossi2019verifai,vardhan2023constrained,vaishya2020artificial} and a better understanding and insight about the model's performance would be useful for practical control designer. Since Deep learning is a data hungry model, it is not possible to generate an enormous amount of data for each application. In such cases, this DNN models fail to perform and alternative learning model can be useful.

\section{Conclusion and Future Work}
\label{sec:conclusionFutureWork}

\subsection{Conclusion}
 In this work, we attempted to design a controller for the lateral control of a vehicle using AI-ML models. Experiments' results show that:
 \begin{enumerate}
     \item  Random forest-based controller has better generalization capability than Deep neural network when limited data is available.
     \item Trained on one track and deployed on another track, the RF-based controller was able to complete the track without crashing, while the deep neural network-based controller  could not complete even 10\% of the track before crashing.
     \item Due to an ensemble of trees in RF, it is possible to quantify uncertainty and can help to decide the failure probability of the controller and take command of control if needed.  
 \end{enumerate}

\subsection{Future Work}
Our experiments are use cases to understand deeper about two prominently used AI models when used in the context of control. For future work, we want to extend this work by replacing the deep neural network with a Bayesian neural network that has better generalization capability and can predict confidence in prediction and evaluate its performance.

\bibliographystyle{IEEEtran}
\bibliography{references}
\end{document}